\documentclass[11pt,a4paper]{article}
\usepackage[hyperref]{emnlp-ijcnlp-2019}

\usepackage{times}
\usepackage{latexsym}
\usepackage{microtype}
\usepackage{graphicx}
\usepackage{subfigure}
\usepackage{booktabs} % for professional tables
\usepackage{latexsym}
\usepackage{multirow}
\usepackage{url}
\usepackage{amsmath}
\usepackage{wrapfig, blindtext}
\usepackage{float}
\usepackage{booktabs}
\usepackage{amssymb}
\usepackage{amsmath}
\usepackage{xcolor}

\usepackage{mathtools}
\usepackage{capt-of}
\usepackage{cleveref}
\usepackage{url}
\usepackage{algorithm}
\usepackage{algorithmic}

\definecolor{myred}{rgb}{0.67,0.1,0.1}

\aclfinalcopy % Uncomment this line for the final submission

\title{Building an Application Independent Natural Language Interface}

\author{Sahisnu Mazumder,~ Bing Liu,~ Shuai Wang,~ Sepideh Esmaeilpour \\
  Department of Computer Science, University of Illinois at Chicago, USA \\
  \texttt{sahisnumazumder@gmail.com,liub@uic.edu}\\
  \texttt{shuaiwanghk@gmail.com,sesmae2@uic.edu} \\}

\date{}

\begin{document}
\maketitle

{\large \color{myred}  An \textbf{extended and revised version} of this work has been published in the HLDS Workshop of NeurIPS 2020 as follows: 
	
	\vspace{2mm}	
	\begin{quote}
		Sahisnu Mazumder, Bing Liu, Shuai Wang, and Sepideh Esmaeilpour. \textbf{An Application-Independent Approach to Building Task-Oriented Chatbots with Interactive Continual Learning}. In NeurIPS-2020 Workshop on Human in the Loop Dialogue Systems, 2020.
		
		\vspace{1mm}	
		HLDS@NeurIPS: {\color{blue} \href{https://sites.google.com/view/hlds-2020/home}{main paper \& supplementary}}
		
		click here for the pdf of main paper: {\color{blue} \href{https://drive.google.com/file/d/11NTjKfCRvEPu1__CPOrwJ1QYlBrNrjIJ/view}{HLDS@NeurIPS 2020 version link}}
	\end{quote}		
	
	\noindent
	\underline{\textbf{Please consider the HLDS@NeurIPS 2020}}\\ \underline{\textbf{version for citation as mentioned above.}} \\
}

\begin{abstract}
Traditional approaches to building natural language (NL) interfaces typically use a semantic parser to parse the user command and convert it to a logical form, which is then translated to an executable action in an application.~However, it is still challenging for a semantic parser to correctly parse natural language. For a different domain, the parser may need to be retrained or tuned, and a new translator also needs to be written to convert the logical forms to executable actions. In this work, we propose a novel and \textit{application independent} approach to building NL interfaces that does not need a semantic parser or a translator. It is based on natural language to natural language matching and learning, where the representation of each action and each user command are both in natural language. To perform a user intended action, the system only needs to match the user command with the correct action representation, and then execute the corresponding action. The system also interactively learns new (paraphrased) commands for actions to expand the action representations over time. Our experimental results show the effectiveness of the proposed approach. 
\end{abstract}

\section{Introduction}
Existing techniques \cite{zelle1996learning,artzi2013weakly,andreas2015alignment,zettlemoyer2012learning,DBLP:series/synthesis/2018Li} for building natural language interfaces (NLIs) often use a semantic parser to parse the natural language (NL) command from the user and convert it to a logical form (LF) and then, translate LF into an executable action in the application. This approach has several limitations: \textbf{(1)} it is still very challenging for a semantic parser to correctly parse natural language.  \textbf{(2)}
For different applications, the parser may need to be retrained/tuned with an application domain corpus. \textbf{(3)} For each application, a different translator is needed to convert the logic forms into executable actions. \textbf{(4)} Due to the last two limitations, it is difficult to build an application-independent system.

This work proposes a novel approach to NLI design, called \textit{Natural Language to Natural Language (NL2NL)}. This approach works in the following setting: Let the underlying application has a finite set of executable actions/functions that the user can use to accomplish his/her goal. These actions or functions for end users are commonly defined by the application as Application Programming Interfaces (APIs). Most applications work in this setting as providing APIs is a standard approach. 
For each API, the proposed approach attaches a natural language representation of it, which is a set of one or more \textbf{\textit{API seed commands}} (\textbf{ASCs}) written in natural language (e.g., by the application developer) just like a natural language command from the user to invoke the API. The only difference is that the objects to be acted upon in each ASC are replaced with variables, which indicate the arguments of the API. These variables will be instantiated to actual objects/arguments in a real user command. Let's see an example. 

Consider an application like \textit{Microsoft Paint} and an API like \texttt{drawCircle($X1$,$X2$)} (drawing a circle having color $X1$ at coordinate $X2$). One ASC for this API can be ``\textit{draw a $X1$ circle at $X2$}". Clearly, $X1$ and $X2$ are arguments of the API and represented as variables in the ASC. When the user gives a natural language command, the system simply matches the command with one of the ASCs and in doing so, also instantiates the objects (i.e., API arguments referred in the user command) for the associated API to be executed. For example, the user command ``\textit{draw a blue circle at (20, 40)}'' can be grounded to the aforementioned ASC, where the grounded API arguments are  $X1=$`\textit{blue}' and $X2=$\textit{(20, 40)}. Since both the user command and the ASCs are written in natural language, we call this approach \textit{natural language to natural language} (\textbf{NL2NL}). 

Since the user may use many different language expressions to express the same command, the matching process is still challenging. For example, given the ASC (`\textit{draw a $X1$ circle at $X2$}'), the user may say ``\textit{insert a circle with color blue at (20, 40)}." The matching algorithm may not be able to match them. Also, if the user says ``\textit{move the circle with color blue to (20, 40)}", it should be grounded/matched to a different API [e.g., \texttt{moveCircle}($X1$, $X2$)]. %Language matching plays a key role to resolve the ambiguity. 
As the developer has to provide API seed commands (ASCs) for the APIs of the underlying application, it can be hard for the developer to write all possible paraphrased commands for a given API. This affects the coverage of the proposed NL2NL system in grounding user commands. To deal with this issue, we also propose an interactive learning mechanism to interact with the end user in natural language to learn new (paraphrased) commands and convert them to new ASCs and add them to the existing set of ASCs (the set enlarges) so that when similar commands are issued by this or other users in the future, the NL2NL approach can handle it. This enables continuous learning of new ASCs from users to improve the subsequent grounding performance.

The proposed system based on NL2NL is called \textbf{CML} (\textit{Command Matching and Learning}). CML has three key advantages: \textbf{(1)} Due to the NL2NL matching, we no longer need a semantic parser to convert the user command to a logical form (or action formalism) or to write a program to translate the logical form to an API call. This makes the design of natural language interfaces much simpler and quick because ASCs are written in natural language (NL) just like user commands, and can be easily written by the application developer for the APIs of their application. \textbf{(2)} Again, as ASCs are written in NL, the matching algorithm of CML can be used for any application and thus is \textit{application-independent}. CML simply maps a user command to a correct ASC, and the system executes the API attached to the ASC. \textbf{(3)} CML also learns new ASCs in the process of being used to make it more powerful. To our knowledge, no existing approach is NL2NL, application independent, or able to learn after deployment.

We evaluate the proposed system CML using two representative applications: (1) \textit{Blocks-World}, and (2) \textit{Webpage Design}. Experimental results show its effectiveness.

\section{Related Work}
Many existing works have been done on building natural language interfaces (NLI) for various applications. 
For robot navigation, \cite{artzi2013weakly} proposed a weakly supervised learning method to train semantic parsers for grounding navigation instructions. \cite{tellex2011understanding} proposed a related method also for navigation and mobile manipulation. \cite{janner2018representation} worked on understanding spatial references in NL for NLIs. \cite{guu2017language} learns a reinforcement learning (RL) based semantic parser with indirect supervision. \cite{andreas2015alignment} gave a sequence-prediction based model for following NL instructions. \cite{fried2017unified} proposed an unified pragmatic model for instruction following. Other prominent works include NLI for scrutable robots \cite{garcia2018explain} and dialog agents for mobile robots to understand instructions through semantic parsing \cite{thomason2015learning}.

%\textbf{Querying Databases.} 
In NLI for databases, \cite{zelle1996learning} used inductive logic programming to construct an NLI for database querying. \cite{zettlemoyer2007online} learns a weighted combinatory categorial grammar (CCG) for flight database queries. \cite{berant2013semantic,yih2015semantic} proposed a semantic parser for question-answering using a knowledge base. Other prominent works on database querying are \cite{baik2019bridging,xiong2019transferable,neelakantan2016learning,li2019spatialnli,ferre2017sparklis,liang2016learning,zhong2017seq2sql} and data exploration and visual analysis \cite{setlur2016eviza,utama2018end,lawrence2016nlmaps,gao2015datatone}. More information can be found in~\cite{DBLP:series/synthesis/2018Li}.

%\textbf{Webpages and Graphical User Interfaces.}  
For webpages and GUIs, \cite{branavan2010reading} proposed a RL-based solution for mapping high-level instructions to API calls. \cite{su2017building} proposed an end-to-end approach to designing NLI for web APIs. A similar approach is also used in \cite{pasupat2015compositional} for performing computations on web tables, \cite{Soh2017tagUI,pasupat2018mapping} designed an NLI for submitting web forms and interacting with webpages and \cite{lin2018nl2bash} proposed an NLI for Bash commands. 

Besides these, there are works on selecting correct objects referenced in utterances~\cite{frank2012predicting,golland2010game,smith2013learning,celikyilmaz2014resolving}, learning language games \cite{wang2016learning} and discovering commands in multimodal interfaces \cite{srinivasan2019discovering}.

All these approaches differ substantially from our work as they are based on learning semantic parsers of various kinds or end-to-end models with labeled examples. % to generate executable script directly from NL command. 
CML is mainly based on natural language to natural language (NL2NL) matching. %approach is for NLI design, which can support both supervised and unsupervised language matching technique to build NLI. 
Due to NL2NL matching, our approach presents an application-independent solution to NLI, in the sense that the application developer does not need to collect application-specific training examples to learn a parser and our matcher can work with any application. We only require the developer to write ASCs in natural language for their APIs, which is much easier to do than collecting labeled data and training a parser or an end-to-end model. Furthermore, CML can learn new ASCs in the usage process (after the system is deployed) to make it more and more powerful.  %Our system does not do state tracking and robot planning.

\section{Proposed Approach}
The proposed approach CML consists of four parts: (1) an \textit{ASC (API seed command) specification}, (2) a \textit{utility constraint marker}, (3) a \textit{command grounding module}, and (4) an interactive \textit{ASC learner}. 
\textit{ASC specification} enables the application developer to specify a set of ASCs for each API of their application. The \textit{utility constraint marker} identifies some sub-expressions in each ASC that utility ASCs should not be applied to. The \textit{command grounding module} grounds a user command to an action ASC (both are in natural language) for an API call. %and then, calls the API to execute the (intended) action. 
The \textit{ASC learner} learns new ASCs from the user to make the system more powerful. An natural language interface (NLI) is then built with no application specific programming or data collection required. % The details of the three parts are given in the next few subsections.

\subsection{ASC Specification}
Since ASCs are written in natural language, they can be specified by the application developer (without knowing how CML works). To achieve the goal of application-independent NLI design, the proposed CML has to automatically read and understand the ASCs for each application. We define an \textit{ASC specification} for developers that can be followed to easily write ASCs for their applications. 

The ASC specification consists of three parts: (1) \textit{properties and value domains} of the application, (2) \textit{action ASC specification} and (3) \textit{utility ASC specification}.
Let the set of actions that can be performed in the application be $\mathcal{A}$. Each action $a_i \in \mathcal{A}$ causes a change in the \textit{state} of the (instantiated) objects in the application, specified by the object properties and their values. For example, in the \textit{Microsoft Paint} application, a circle drawn on the editor is an example of \textit{instantiated} object and it can have properties like \textit{color}, \textit{name}, \textit{shape}, etc, with their values like the color of the circle being \textit{red}. And examples of actions are: \textit{draw a circle}, \textit{change the shape of the circle to make it a square}, etc. Each of these actions is performed by an unique API call, referred to as an \textbf{action API}. The developer defines one or more action ASCs [in (paraphrased) natural language] for each action API, where the arguments of the API are variables in the corresponding action ASC (see the Introduction section).

Besides action APIs, there are also \textbf{Utility APIs}, which are used to retrieve information about the properties of the \textit{\textbf{instantiated}} object and their values from the current application state and are used as helper functions to $\mathcal{A}$. The concept of \textbf{Utility API} is explained as follows. 

Often a natural language command from the user involves one or more \textit{referential expressions} to \textit{instantiated} objects, which need to be resolved to interpret the user command. A \textit{referential expression} is a phrase used to refer an object \textit{indirectly} by a property value. For example, in the command: ``\textit{move the blue block to the left of the cube}", `blue block", ``cube", ``left of the cube'' are \textit{referential expressions}, where `\textit{blue block}" and ``\textit{cube}" refers to some blocks having color ``\textit{blue}" and shape ``\textit{cube}" respectively. The phrase ``\textit{left of the cube}'' refers to some \textit{location} in the application space, where the blue block is to be moved to. 

Since there may not be any action API that consider such indirect object references \textit{as a part of its arguments}, the aforementioned user command cannot be \textit{directly} grounded to any of the action APIs. In such cases, utility APIs are used to resolve the \textit{referential expressions} by identifying the referred objects, having a given property and a value. Thus, unlike action APIs, a utility API returns one or more values as output (e.g., object ids) to the API call which is used by an action API and/or other utility APIs to unambiguously ground the user command. Similar to action ASCs, the developer also defines one or more utility ASCs [in (paraphrased) natural language] for each utility API, where the arguments of a utility API becomes variables in the corresponding utility ASC. 

%Apart from specifying action and utility ASCs, the developer also defines \textit{utility usage constraints} or simply, \textbf{\textit{utility constraints}} for each action ASC to restrict the use of some utility APIs for that action. We will discuss these constraints in the ``Command Matcher of CML" subsection.

\begin{table}[t!]
\vspace{-0.3cm}
\small
\centering
\caption{\small Properties and their domains for Blocks-World}
\begin{tabular}{|l|l|}
\hline
\multicolumn{1}{|c|}{\textbf{Property}} & \multicolumn{1}{c|}{\textbf{Domain}}                                                                 \\ \hline
color  (object)                                 & \begin{tabular}[c]{@{}l@{}}'red', 'green', 'orange',\\  'blue', 'yellow'\end{tabular}                \\ %\hline
shape (object)                                & \begin{tabular}[c]{@{}l@{}}'triangular', 'circular', 'cube',\\  'square', 'rectangular'\end{tabular} \\ \hline
location (object)                                 & (x, y) coordinate in 2D space                                                                        \\ %\hline
name (object)                                    & English alphabets A-Z                                                                                \\ %\hline
direction (action)                              & 'right', 'left', 'above', 'below'                                                                    \\ \hline
\end{tabular}
\vspace{-0.35cm}
\end{table}

\begin{table*}[t!]
\scriptsize
\centering
\caption{\small Action ASC specifications for Blocks-World and groundable example NL commands from user. (*) denotes that the variable do not take part in command reduction (Utility Constraints), which is automatically detected and marked by CML. (X denotes input)}
\begin{tabular}{|p{2.25cm}|c|p{2.9cm}|p{3.75cm}|p{4.5cm}|}
\hline
\multicolumn{1}{|c|}{\textbf{API Function}}                            & \multicolumn{1}{l|}{\textbf{AID}} & \multicolumn{1}{c|}{\textbf{Action ASCs}}                                 & \multicolumn{1}{c|}{\textbf{Variable/Argument: Type}}                                                        & \multicolumn{1}{c|}{\textbf{Example User Commands}}                                                    \\ \hline
Add(X1, X2)                                                        & \textbf{1}                        & \begin{tabular}[c]{@{}l@{}}add a block at X1;\\ insert a block at X1\end{tabular}        & 'X1': 'location'(*)                    & add a block at row 2 and column 3; put a block at (2, 3)         \\ %\hline
Remove(X1)                                                     & \textbf{2}                        & remove X1                                                               & 'X1': 'block\_set'                                                                                  & \begin{tabular}[c]{@{}l@{}}delete blue block; take away blue\end{tabular}                                 \\ %\hline
Move(X1, X2)                                                       & \textbf{3}                        & \begin{tabular}[p{2.77cm}]{@{}l@{}}move X1 to X2; shift X1 to X2\end{tabular}       & 'X1': 'block\_set',  'X2': 'location'(*)                    & \begin{tabular}[c]{@{}l@{}}move blue block to the left of cube; shift green \\cube to (4, 5)\end{tabular}   \\ %\hline
\begin{tabular}[c]{@{}l@{}}MoveByUnits(X1, \\ X2, X3)\end{tabular} & \textbf{4}                        & \begin{tabular}[c]{@{}l@{}}move X1 along X2 \\ by X3 units\end{tabular} & \begin{tabular}[c]{@{}l@{}}'X1': 'block\_set', 'X2': 'direction', \\ 'X3': 'number'\end{tabular} & \begin{tabular}[c]{@{}l@{}} move blue block left by 2 units;~ shift green \\ cube down by 3 units \end{tabular}\\ %\hline
UpdateColor(X1, X2)                                                & \textbf{5}                        & change color of X1 to X2; color X1 X2   & 'X1':'block\_set', 'X2':'color'(*)                       & color A red; change color of B to blue                           \\ %\hline
UpdateShape(X1, X2)                                                & \textbf{6}                        & change shape of X1 to X2           & 'X1':'block\_set',  'X2':'shape'(*)                       & \begin{tabular}[c]{@{}l@{}}set the shape of A to cube; make B square\end{tabular}                         \\ %\hline
Rename(X1, X2)                                                     & \textbf{7}                        & rename block X1 to X2           & 'X1': 'block\_set',  'X2':'name'(*)                        & Name the block at (4, 5) as C; rename A to D                      \\ \hline
\end{tabular}
\end{table*}

\begin{table*}[t!]
\scriptsize
\centering
\caption{\small Utility ASC specifications for Blocks-World and groundable example sub-expressions (O denotes output, X denotes input)}
\begin{tabular}{|p{2.4cm}|c|p{1.5cm}|p{3.5cm}|p{4.25cm}|}
\hline
\multicolumn{1}{|c|}{\textbf{API Function}} & \textbf{AID} & \multicolumn{1}{c|}{\textbf{Utility ASC}}                         & \multicolumn{1}{c|}{\textbf{Argument: Type}}                                                          & \textbf{Example sub-expressions}                                                                           \\ \hline
GetBlocksbyColor(X1)                     & \textbf{8}   & color/X1                                                               & X1: 'color',  O1: 'block\_set'                              & blue block                                                                                              \\ %\hline
GetBlocksbyShape(X1)                     & \textbf{9}   & shape/X1                           & X1: 'shape', O1: 'block\_set'                               & square block; cube                                                                                      \\ %\hline
GetBlocksbyName(X1)                      & \textbf{10}  & name/X1                                                                & X1: 'name', O1: 'block\_set'                                & block A                                                                                                 \\ %\hline
GetBlocksbyLocation(X1)                  & \textbf{11}  & location/X1                                                            & X1: 'location', O1: 'block\_set'                           & block at row 2 and  column 3; block at (2, 3)              \\ %\hline
GetLocation(X1, X2)                     & \textbf{12}  & \begin{tabular}[c]{@{}l@{}}get location \\ along X1 of X2\end{tabular} & \begin{tabular}[c]{@{}l@{}}'X1': 'direction', 'X2': 'block\_set', \\ 'O1': 'location'\end{tabular} & \begin{tabular}[l]{@{}l@{}}at the left of blue block;\\below block B\end{tabular} \\ \hline
\end{tabular}
\vspace{-0.2cm}
\end{table*}

\subsection{Blocks-World Specification: An Example}
We now give an example specification for a Blocks-World application, which is about arranging different blocks or tiles on a 2D grid or adding them to form a goal arrangement. % specified in a goal command provided by the application to the user. 
Note that, our CML system is not concerned with the goal or how to achieve it, which is the responsibility of the user. CML only matches each user command utterance to an action ASC for the associated action API to be executed in the environment, with the help of utility ASCs. 

In blocks-world, the objects are basically tiles of different \textit{shapes}, \textit{colors} and \textit{names} and the action APIs are designed to manipulate these objects in the 2D space. The utility APIs retrieve values of the properties like \textit{shape}, \textit{color} and \textit{name}, etc., of an instantiated tile. We use the term \textit{block} and \textit{tile} interchangeably on-wards.

\subsubsection{Properties and Domains} Table 1 shows different properties of an object or action (specified in the bracket next to the property name). The domain column in Table 1 lists the domain of each property. For example, ``\textit{color}" is a property of a tile, which can take one of five possible values \{`\textit{red}', `\textit{green}', `\textit{orange}', `\textit{blue}', `yellow'\}. Similarly, \textit{shape} denotes the shape of a tile and can be one of five categories:
\{`\textit{triangular}', `\textit{circular}', `\textit{cube}', `\textit{square}', `\textit{rectangular}'\}. The \textit{name} of a block is denoted by any letter from A-Z. The \textit{location} of an object is specified by a 2D co-ordinate ($x$, $y$) and so on. Besides these, `\textit{direction}' is a property with domain \{`\textit{left}', `\textit{right}', `\textit{above}', `\textit{below}'\} needed to execute an action for moving an object in a given direction. 

\subsubsection{ASC Specification}
Table 2 shows the action ASCs and Table 3 shows the utility ASCs and also, some example user commands and sub-expressions (explained later) that can fire them. Note that, all ASCs are written in natural language and there is no fixed format. Here, the argument type `\textit{block\_set}' denotes a set of unique (instantiated) block object ids. We define seven action ASCs and five utility ASCs to ground a user command for Blocks-World. The action ASCs are: adding a block at location ($x$, $y$) [AID 1], removing a block [AID 2], moving blocks [AID 3-4], changing properties of a block [AID 5-7]. Utility ASCs either retrieve the \textit{block\_set} for a given input property value [AID 8-11] or returns a location object ($x$, $y$) in a given direction of a block [AID 12]. 
The arguments marked (*) in Table 2 denote the utility constraints automatically identified by CML for restricting the use of utility ASCs, which we discuss next. 

% \noindent
\subsection{Utility Constraint Marker} 
For some actions, some variables (arguments) in the user command should not be reduced as it can result in incorrect subsequent grounding. As utility ASCs are used to resolve sub-expressions in a user command, if a sub-expression (a sub-sequence of user command, defined in next subsection) matches both \textit{partially} with an action ASC and \textit{fully} with a utility ASC, it creates ambiguity (see below). A \textbf{\textit{sub-expression is said to be matched}} with an ASC, if the sequence of argument types of the variables appearing in the sub-expression from left to right also appears in the ASC. And by \textit{full} match, we mean the number of argument types matched is equal to the number of argument types present in the ASC. We mark each such argument appearing in the \textit{action ASC} with a * for the action ASC indicating a \textit{utility constraint}. Note, if a sub-expression matches \textit{fully} with an action ASC, there is no ambiguity as the corresponding action API will be automatically fired. 

Let's have an example: Consider the ASC with AID 1 in Table 2 for adding a block at location X1. Here, the argument \textit{X1:location} [marked (*)] should not be resolved using utility ASC of AID 11 (getting the block(s) at the location). Otherwise, it will mislead the grounding by reducing a user command like ``add a block at (2, 3)" to ``add a block at block\_set/X1". Similar problems arise when CML attempts to resolve arguments like color ($X2$) in ASC with AID 5, shape ($X2$) in ASC with AID 6, etc., using utility ASCs with AID 8 and AID 9 respectively, while grounding user commands like ``color block A to \textbf{red}'' and ``make the blue block \textbf{square}" respectively. we use * to mark these action ASC arguments to indicate no reduction should be applied. 

Formally, 
let $u_{seq}= \langle type(X1_u), . . ., type(XN_u) \rangle$ be the sequence of variable types in a utility ASC $u$ from left to right, where $Xi_u$ is the $i^{th}$ variable in $u$. Similarly, let $a_{seq}$ be the sequence of variable types (from left to right) in an action ASC $a$. Let $M_{seq}$ be the longest common sub-sequence of $u_{seq}$ and $a_{seq}$. Then if $|M_{seq}| = |u_{seq}|$, all variables corresponding to the argument types in $M_{seq}$ should be marked with (*) to indicate utility constraints for action ASC $a$.

% While reading the ASC specification $S$, CML uses the aforementioned rule to automatically mark the variables in each action ASC with utility constraint markers, before it begins processing of any user command.

\subsection{Command Grounding Module of CML}
Given a natural language command $C$ from the user and the ASC specification $S$ for an application, the \textbf{\textit{command grounding module}} (CGM) returns a grounded ASC tuple $\hat{A}$, consisting of one \textit{action} ASC and a set of \textit{utility} ASCs. They together tell the system what action (API call) to perform. If the grounding is not possible, $\emptyset$ is returned.  

CGM consists of two primary modules: a \textbf{\textit{tagging and rephrasing module}} ($\mathcal{R}$) and an \textbf{\textit{ASC matching module}} [or simply  \textbf{\textit{Matcher}}] ($\mathcal{M}$). $\mathcal{R}$ tags the input user command $C$ and then, repharses the tagged $C$ to get a command $C'$ that is lexically closer to the ASCs. Here, a tag is an argument type of an action ASC. For example, in command "\textit{move the cube to (2,3)}", the phrase ``\textit{cube}" is tagged as ``shape" and (2, 3) is tagged as ``location" for blocks-world. This work uses a dictionary lookup based and regular expression-based tagger for $\mathcal{R}$\footnote{As most NLI applications have finite domains of objects and properties, we found simple lookup based tagging works well. In complex scenarios, $\mathcal{R}$ can be learned through user feedback or provided by the developer.}. While reading the specification $S$, $\mathcal{R}$ forms a tagset by enumerating all \textit{argument types} in action ASCs and also, builds a vocabulary $V$ consisting of all the words in the ASCs. Given $C$, $\mathcal{R}$ either maps individual words or phrases in $C$ into one of the tags, or repharses them by replacing synonym words/phrases from $V$ using WordNet \cite{miller1990introduction} and ConceptNet \cite{speer2017conceptnet}.

Given the rephrased and tagged user command $C'$ and the set $\mathcal{T}$ of (action or utility) ASCs for an application, Matcher $\mathcal{M}$ computes a match score $s$($t$, $C'$) for each $t \in \mathcal{T}$ and returns the top ranked ASC~$\hat{t} = \arg\max_{t~\in~\mathcal{T}} s(t, C')$. Any paraphrasing model can be used as $\mathcal{M}$. This work uses information retrieval (IR) based unsupervised matching models for $\mathcal{M}$ (we compare different types of unsupervised IR matching models in the Experiment section).

\subsubsection{Working of Command Grounding Module (CGM)} Algorithm 1 shows the iterative grounding process of a user command $C$ by CGM. We again use the Blocks-World application (see Figure 1) to explain the algorithm on-wards.

% as \textbf{\textit{utility constraints for that action ASC}} (see * in Table 2).

\setlength{\textfloatsep}{0.2cm}
\setlength{\floatsep}{0.2cm}
\begin{algorithm}[tb]
	\small
	\caption{\small Iterative Command Grounding}
	\label{alg:example}
	\begin{flushleft}
		\textbf{Input:} $C$: natural language command issued by user;\\
		\hspace{0.85cm} $\mathcal{T}$: ASC store;\\
		\hspace{0.85cm} $\mathcal{R}$: Command tagger and rephraser; \\
		\hspace{0.85cm} $\mathcal{M}$: ASC Matcher;\\
		\textbf{Output:} $\hat{A}$: predicted AID set for grounding $C$;
	\end{flushleft}
	\vspace{0.1cm}
	\begin{algorithmic}[1]
	    \STATE $C'~\leftarrow$ \textbf{\texttt{Semantic\_Tagging\_Rephrasing}}($C$, $\mathcal{R}$)  \COMMENT{\small{$C'$ is the tagged and rephrased user command}}
	    \STATE $SP~\leftarrow$ \textbf{\texttt{Enumerate\_Filter\_Subexpressions}}($C'$, $\mathcal{T}$, $\mathcal{M}$)  \COMMENT{\small{$SP$ is the list of enumerated and filtered sub-expr. in $C'$}}
	    \STATE $\hat{A}~\leftarrow~\emptyset$;$~~j~\leftarrow$ 0
	    
	    \vspace{0.1cm}
		\WHILE{$TRUE$}  %
		\STATE $A_{set}~\leftarrow$ \textbf{\texttt{GetCandidateASCs}}($\mathcal{T}$, $C'$, ``\textit{\textbf{action}}")  %\COMMENT{Retrieve}
		\IF{$A_{set} = \emptyset$}
		\IF{$SP = \emptyset$ ~or~ $j > |SP|-1$}
		\RETURN $\emptyset$
		\ELSE
		\STATE $U_{set}~\leftarrow$ \textbf{\texttt{GetCandidateASCs}}($\mathcal{T}$, $SP_j$, ``\textit{\textbf{utility}}")  \COMMENT{$SP_j$ is the $j^{th}$ sub-expression in $SP$} %\COMMENT{Retrieve}
		\IF{$U_{set} \neq \emptyset$}
		\STATE $u_r ~\leftarrow$ \textbf{\texttt{GetTopRankedASC}}($U_{set}$, $SP_j$, $\mathcal{M}$)  %\COMMENT{Retrieve}
		\STATE $\hat{A}~\leftarrow~ \hat{A}~ \cup~ \{u_r\}$
		\STATE $C'~\leftarrow$ \textbf{\texttt{ReduceCommand}}($C'$, $SP_j$, $u_r.output$)
		\STATE $SP~\leftarrow$ \textbf{\texttt{Enumerate\_Filter\_Subexpressions}}\\($C'$, $\mathcal{T}$, $\mathcal{M}$)
		
		\STATE $~~j~\leftarrow$ 0
		\ENDIF
		\STATE $j~\leftarrow~j+1$
		\ENDIF
		\ELSE
		\STATE $a_k~\leftarrow$ \textbf{\texttt{GetTopRankedASC}}($A_{set}$, $C'$, $\mathcal{M}$)
		\RETURN $\hat{A}~ \cup~\{a_k\}$
		\ENDIF
		\ENDWHILE
	\end{algorithmic}
	\normalsize
\end{algorithm}
\setlength{\textfloatsep}{0.4cm}
\setlength{\floatsep}{0.4cm}

% \noindent
$\bullet~$ CGM first performs tagging and rephrasing of $C$ using $\mathcal{R}$ (Line 1). For example, the user command $C$ in Figure 1 is tagged to "\textit{relocate the \textbf{color/X1} block to the \textbf{direction/X2} of \textbf{name/X3}}" [where, \textit{X1}="\textit{blue}", \textit{X2}="\textit{left}", \textit{X3}="\textit{D}"] and rephrased command is $C'_0$ in Figure 1 ($C'_0$ is $C'$ in Line 1 of Algorithm 1 at the moment), where ``\textit{relocate}" is replaced with synonym ``\textit{move}" found in vocabulary $V$.

% \noindent
$\bullet~$ Next, CGM enumerates and extracts the list of sub-expressions ($SP$) present in $C'$ (Line 2) to assist compositional grounding of $C'$ using utility ASCs in subsequent steps of the algorithm. 
A sub-expression of length-$m$ is defined as a substring of $C'$ involving $m$ consecutive variables (with types) and intermediate words linking them. For example, given $C'_0$ in Figure 1, the list of all sub-expressions: $SP$ = [ ``\textit{color/X1}", ``\textit{direction/X2}", ``\textit{name/X3}", ``\textit{color/X1 block to the direction/X2}", ``\textit{direction/X2 of name/X3}"], where the first three in $SP$ are length-1 and last two are length-2 sub-expressions. Length-$3$ sub-expression is the full command $C'_0$ and is discarded from $SP$ as it does not take part in matching the utility ASCs, only matched with action ASCs.

Here we also filter out those sub-expressions that are marked with (*) (with utility constraints) as they should not be reduced by utility ASCs. For this, CGM first uses Matcher $\mathcal{M}$ to identify the top-ranked action ASC semantically close to $C'$ (e.g., AID 5 for the user command ``color block A to \textbf{red}") and then, delete all sub-expressions involving such arguments (e.g., ``\textit{color/X2}") from $SP$ [Line 2].

With the resulting $SP$, the ASC grounding for $C'$ [Lines 4-24] happens as follows:

$\bullet~$ First, a candidate set $A_{set}$ of action ASCs is retrieved from the set of action ASCs such that for any $a \in A_{set}$, $C'$ and $a$ has an identical set of variables and their types. If $A_{set} \neq \emptyset$, $\mathcal{M}$ finds and returns the top ranked action ASC $a_k$ [Lines 20-21] for $C'$. In Figure 1, none of the action ASCs are matched directly for $C'_0$ and so, $A_{set} = \emptyset$.

\begin{figure}[t!]
	\centering
	\includegraphics[height=4.32cm]{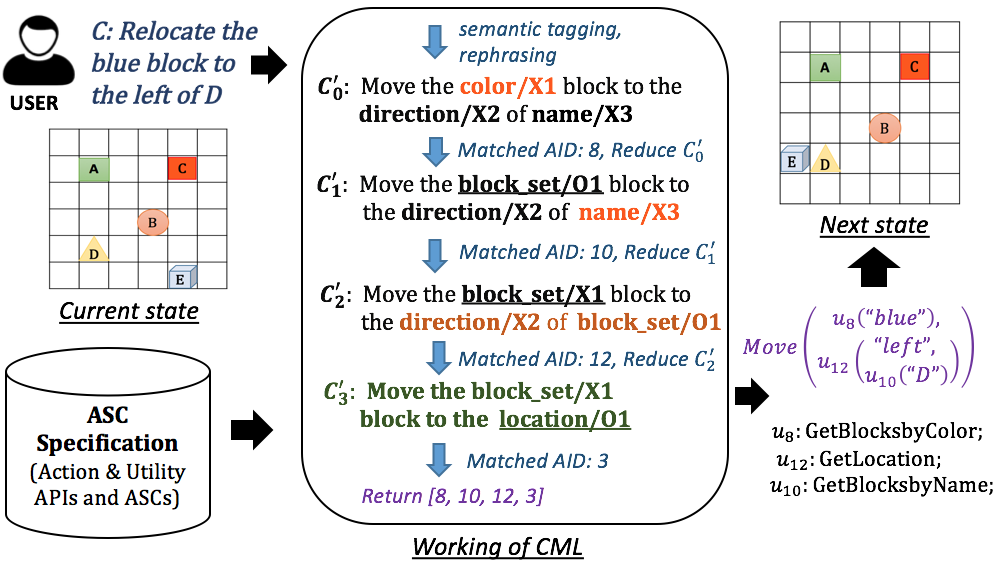}
	\caption{\small An example of working of CGM on a user command for Blocks-World. Here, AID denotes ASC ID (see Tables 2 and 3).}
	\label{task_example}
	\vspace{-0.2cm}
\end{figure}

$\bullet~$ If $A_{set} = \emptyset$, Matcher $\mathcal{M}$ works as follows: If $SP = \emptyset$ or all sub-expressions in $SP$ have been checked, it returns $\emptyset$ indicating grounding of $C$ is not possible [Lines 7-8]. Otherwise, it selects the sub-expressions from $SP$ one by one and then, reduces $C'$ further by resolving bindings for the variables present in that sub-expression [Lines 10-18]:
Given the $j^{th}$ sub-expression $SP_j$, Matcher first selects a candidate set $U_{set}$ of utility ASCs from $\mathcal{T}$ such that for any $u \in U_{set}$, $SP_j$ and $u$ has identical set of variables and their types [Line 10]. Note, while matching the variables with utility ASCs, we \textbf{rename} the variables in $SP_j$ in order to avoid error in matching due to different variable names.  For example, variable X3 in sub-expression ``\textit{name/X3}" [see $C'_1$ in Figure 1] is renamed as X1 [i.e.``\textit{name/X1}"] (not shown), so that it can match with AID 10, while reducing $C'_1$ to $C'_2$.
Similarly, X2 and O1 [in $C'_2$ of Figure 1] in sub-expression ``\textit{direction/X2 of block\_set/O1}" are renamed as X1 and X2 [i.e., ``\textit{direction/X1 of block\_set/X2}"] so that it can match with AID 12.%, while reducing $C'_2$ to $C'_3$.

$\bullet~$ If $U_{set} = \emptyset$, Matcher $\mathcal{M}$ cannot perform reduction of $C'$ for $SP_j$ and only $j$ is incremented [Line 18]. Otherwise, $\mathcal{M}$ returns the top ranked utility ASC $u_r$ for $SP_j$ [Line 12]. 
Next, $C'$ is reduced by replacing $SP_j$ with "\textit{type}(O1)/O1" [the output variable and its type corresponding to $u_r$] (Line 14). For example, given $SP_j$ =``\textit{name/X3}" [see $C'_1$ in Figure 1], utility AID 10 gets matched and $C'_1$ gets reduced to $C'_2$, where ``O1" is the output variable and $type(O1)$= ``\textit{block\_set}" [i.e., set of block ids]. As a part of reduction, the variables of $C'$ obtained after replacement are also renamed [i.e. from \textit{block\_set/O1} in $C'_2$ in Figure 1 to \textit{block\_set/X3} (not shown)] and aliases are recorded, so that in the next iteration, it can be matched with the action ASCs [in Lines 5 and 21]. After reduction, we again enumerate and filter $SP$ using the new (reduced) $C'$ and set $j$ to 0 (Line 15-16) similar to Line 2. CGM also stores the values of the output variables in a buffer obtained by executing the functions of the matched utility ASCs for subsequent grounding. In Figure 1, $C'_3$ matches with action API of AID 3 and the iterative grounding completes here. The final grounded AID list returned by CGM for the example in Figure 1 is $[8,10,12,3]$.

\subsection{ASC Learner of CML}
Usually the developer can provide only a small set of action ASCs for each action API. But the users may come up with many different paraphrased commands for which CML cannot find significant match in the current ASC set provided by the developer, which results in low recall. To deal with this issue, we enable CML to learn new ASCs from users through interactions, as discussed below.

Given a user command $C$ and CML has grounded $C$ into some action ASC/API or the grounding has failed (i.e., Line 8 in Algorithm-1 is executed), it asks user to verify whether the grounding output is correct or not. In this step, the verification question asked to the user is formulated based on fixed natural language (NL) templates (e.g. ``\textit{Am I correct?[yes/No]}" or ``\textit{Do you agree?[yes/No]}") and expects a yes/no answer from the user. If the user says ``yes" or remains silent, the system assumes its prediction is correct and requests next command to be grounded. If the user says ``No" (i.e., prediction is incorrect), CML shows a top-$k$ list of action ASCs (ranked based on match score by $\mathcal{M}$) in natural language (NL, for easy understanding) one for each action API for the user to choose the correct one from (see below). As the number of action APIs in our application is small, we choose $k$ to be the total number of action APIs in the experiments. Thus, the user can choose the correct ASC by scrolling down the list of $k$ options in one go. 

The ranked list of action ASCs in NL are generated as follows. Given a user command $C$ [say, ``\textit{put a block to the left of A}"], CML tags $C$ with $\mathcal{R}$ and iteratively reduces the command to get a final reduced command $C'$ [in this case, ``\textit{put a block to location/X1}", where $X1$=(2, 3) is detected while resolving the reference ``\textit{left of A}" using utility ASCs]. Given the final reduced command $C'$, CML replaces the variables in $C'$ with the corresponding value to get a NL command equivalent to the original user command $C$. In the example, the reduced NL command will be ``\textit{put a block to (2,3)}". Such a NL command is generated for each action ASC, provided there is at least one match in argument values.

If the user chooses one from the ranked list, the learner asks for verifying the correctness of the detected argument values. If the user confirms, ASC learner gathers ground truth action API along with a new ASC and add it into the action ASC set. If the user thinks none of the commands are correct or the intended action is not there in the rank list, user enters a rephrased/rectified version of the command (in a text box), which is again utilized to generate a new rank list of action ASCs. This process repeats until the user chooses an option from the rank list to provide the ground truth or maximum $m$ attempts are made (in which case, the learner learns no new ASC for the current interaction process).

\section{Experiments}
\vspace{-0.1cm}
We evaluate CML on two representative applications: (1) \textit{Block-World} and (2) \textit{Webpage Design}.\footnote{In terms of user commands, many applications are similar to our experimental applications, e.g., building graphical user interfaces, robot navigation, drone control, etc. This initial work does not handle compositions of actions which is required by some complex tasks such as database queries. We plan to do it next.} To create the \textbf{test data} for each application, we showed the supported API functions 
% and some ASC examples 
of each application to 
%the ASC Specification with some example NL commands (see Table 2 in paper and Supplementary) for each supported action APIs (to understand the ASCs) to 
five users (graduate students, who are unaware of the working of CML) and asked them to write commands to play with the application to gather the evaluation data (Table 4). We also asked them to write down some commands that are not groundable to any of the action APIs. We randomly shuffle the list of commands and run CML to ground them one by one. Next, we asked the same users to mark the correctness of the grounding results. Based on this, we compute the \textit{accuracy} of various versions of CML. In computing accuracy, we consider the true action API of non-groundable user commands as AID 0, indicating that it cannot be grounded to any of the action ASCs (or APIs) present in the specification. If CML can detect it is non-groundable, it is considered correct. \textbf{ASC specifications} for the Blocks-World applications were given in Tables 2 and 3. The \textbf{ASC specifications} for Webpage Design are given in the Supplementary material. We will release the code and the datasets after paper acceptance. 

\begin{table}[t!]
\small
\centering
\caption{\small Dataset statistics. Here, $m$-UC denotes the number of user commands that needs $m$ utility ASCs for grounding. "Non-Groundable" denotes the number of commands for which no grounding exists for the given API set.}
\begin{tabular}{|c|c|c|c|c|c|}
\hline
0-UC & 1-UC & 2-UC & \textgreater{} 2-UC & \begin{tabular}[c]{@{}c@{}}Non-\\Groundable\end{tabular} & Total \\ \hline
\multicolumn{6}{|c|}{\textbf{Blocks-World}}                   \\ \hline
15     &  160    &   20   &     76     &    42        & 313      \\ \hline
\multicolumn{6}{|c|}{\textbf{Webpage Design}}           \\ \hline
13   &  146    &  13    &     20      &   14         & 206      \\ \hline
\end{tabular}
\end{table}

\subsection{Compared Models} 
As there is no existing work using the NL2NL approach for NLI design, we compare various versions of the CML model for evaluation. Note that we don't compare with existing parsing and/or end-to-end methods as they need application training data and/or are thus not application independent. 

\textbf{(1) CML-jaccard :} This version of CML uses the Jaccard similarity as the scoring function of \textit{Matcher} $\mathcal{M}$. 

\textbf{(2) CML-vsm :}  This version uses tf-idf based vector space model for $\mathcal{M}$ where all ASCs associated with each API are regarded as one document and the user command as query. It is slightly poor if ASCs are treated individually. 

\textbf{(3) CML-emb:} This version uses word embedding based matching model for $\mathcal{M}$. Given an ASC $t$ and a user command (after tagging) $C'$, we retrieve the pre-trained word embedding vectors for each word in $C'$ ($t$) and average them to get the vector representation of $C'$ ($t$) as $v_{I'}$ ($v_t$). Next, we use cosine similarity as the scoring function to measure the similarity between $v_{I'}$ and $v_t$. We use 50D Glove embeddings for evaluation. 

\textbf{(4) CML-vsm (-R):} Variant of CML-vsm, where the rephrasing of words in the input command [Line 1, algorithm 1] is disabled.  

\textbf{(5) CML-vsm (-U):} Variant of CML-vsm, where the use of utility ASCs while grounding [i.e., Lines 7-19, algorithm 1] is disabled.

For (4) and (5), although we only discuss CML-vsm variant in next subsection, we also compared these variants for CML-emb and CML-jaccard as well and found poorer results.  Note also for all CML variants mentioned above, we disabled the use of ASC Learner for learning new ASCs.

\textbf{(6) CML-Y + ASC Learner:} This is the version of CML-Y (where Y is \textit{jaccard}, \textit{vsm}, or \textit{emb}), where we allow CML-Y to learn \textit{new ASCs} through user interactions. Since the emb version is relatively poorer (discussed next), we chose to compare the \textit{jaccard} and \textit{vsm} variants of ASC Learner. 

\begin{table}[t!]
\small
\centering
\caption{\small Overall accuracy comparison of CML variants}
\begin{tabular}{|p{2.5cm}|c|c|}
\hline
\multicolumn{1}{|c|}{\textbf{Compared Models}}                           & \textbf{Blocks-World} & \textbf{\begin{tabular}[c]{@{}c@{}}Webpage Design\end{tabular}} \\ \hline
CML-jaccard                                                              &             \textbf{67.73}    &   80.58                                                                  \\ %\hline
CML-vsm                                                                  &             67.41     &  \textbf{81.06}                                                                       \\ %\hline
CML-emb                                                             &                 \textbf{67.73}      &  77.18                                                                       \\ %\hline
CML-vsm (-R)    &    58.86       &   67.82                                                                      \\ %\hline
CML-vsm (-U)   &    15.97    &    13.10                                                                     \\ \hline
CML-jaccard + & & \\ASC Learner    &    \textbf{68.69}       &   \textbf{83.98}                                                                      \\ %\hline
CML-vsm + & &\\ ASC Learner   &    \textbf{67.73}    &    \textbf{83.01}                                                                     \\ \hline
\end{tabular}
\end{table}

\subsection{Results and Analysis} 
Table 5 shows the accuracy comparison of CML variants on Blocks-World and Webpage Design. CML-jaccard and CML-vsm perform better overall. 
The drop in performance for CML-emb in Webpage Design is primarily due to the out of vocabulary words (no pre-trained embedding is available) in user commands. The performance of CML-vsm(-R) drops significantly, % We observed that, the user often uses diverse paraphrases which makes it hard for the rephraser to map them into the restricted ASC vocabulary. Also, the performance improvement of CML-vsm over CML-vsm(-R) 
which shows that rephrasing helps substantially in command reduction and grounding. CML-vsm(-U) performs the worst among all variants which shows the importance of user command reduction using utility ASCs. % (and APIs).

As CML-jaccard and CML-vsm perform the best overall for both applications, we also compare the performance of ASC Learning variants of these two CML versions in Table 5. We can see that ASC learning clearly improves the performance for both applications. It is very important to note that these improvements are gained from the existing datasets, which do not have many similar commands to the newly learned ASCs. In practice, if similar commands are repeated by many users, the improvement will grow substantially. The performance improvement for Webpage Design is more than for Blocks-World, which can be explained as follows. For Blocks-World, the arguments and their types in API and ASC specifications (see Tables 2 and 3) are quite distinguishable from each other. Thus, correctly identifying arguments and their values in user commands plays a major role in the success of command grounding. Learning of new action ASCs does not make significant impact here. But, for Webpage Design specifications (see Supplementary), action ASCs for APIs 8, 9 and 10 have exactly the same arguments and types, but they differ significantly in action intents. Thus, learning new ASCs helps greatly. 

To investigate the effect of ASC learning further, we evaluate CML-jaccard and CML-jaccard + ASC Learner (CML-vsm and CML-vsm + ASC Learner) on user commands that are only groundable to any of the action APIs with AIDs 8, 9 and 10, as shown in Table 6. Here, we observe almost 8\% improvement in accuracy for ASC Leaner variants of CML, which justifies the explanation.

In Table 7, we compare CML-vsm over various command types (listed in Table 4). Here, we see that, for 0-UC and 1-UC, CML-vsm performs significantly better than for 2-UC, $>$2-UC and NOG (note, for NOG user commands, the true AID is considered as 0) as these commands are harder to ground due to the requirement of multiple (recursive) reduction steps using utility ASCs.

\begin{table}[t!]
\small
\centering
\caption{\small Accuracy improvement of CML ASC Learner variants on user commands groundable to action APIs with AIDs 8, 9 and 10 in Webpage Design (see Supplementary)}
\begin{tabular}{|p{3.77cm}|c|}
\hline
\multicolumn{1}{|c|}{\textbf{Compared Models}}                           & \textbf{\begin{tabular}[c]{@{}c@{}}Webpage Design\end{tabular}} \\ \hline
CML-jaccard   &   82.92                                                                  \\ %\hline
CML-vsm   &  78.04                                                                       \\ \hline

CML-jaccard + ASC Learner    & \textbf{90.24}                                                                      \\ %\hline
CML-vsm + ASC Learner    &    \textbf{85.36}                                                         \\ \hline
\end{tabular}
\end{table}

\begin{table}[]
\small
\centering
\caption{\small  Accuracy for various command categories. Here, NOG denotes ``Non-groundable".}
\begin{tabular}{|l|c|c|c|c|c|}
\hline
\textbf{Application} & \textbf{0-UC} & \textbf{1-UC} & \textbf{2-UC} & \textbf{\textgreater 2-UC} & \textbf{NOG} \\ \hline
Blocks-World   &   53.33   &    85.0  &    40.0    &   59.21   &      33.33             \\ \hline
\begin{tabular}[l]{@{}l@{}}Webpage \\Design\end{tabular}  & 100  &    84.93  &  69.23   & 55.00              &           71.42             \\ \hline
\end{tabular}
\vspace{-0.1cm}
\end{table}

\textbf{Error Analysis.} Overall, we can see that CML variants perform better for Webpage Design than for Blocks-World. On analyzing the datasets, we found that due to the flexibility of the Blocks-World application, the user commands for it are more ambiguous and less specific compared to those in Webpage Design, where users use more application-specific terminologies, which helps grounding. Apart from that, we identified some common grounding errors: \textbf{(1) Argument detection and rephrasing}: failures of CML in detecting argument values in user command or rephrasing due to lack of semantic knowledge in $\mathcal{R}$. E.g., given the command ``\textit{get the ring shaped block out}", CML cannot map the phrase `get out' to `remove' and `ring shaped' to shape 'circular', which cause failure in grounding. \textbf{(2) Ambiguous semantics of words}: In user command  ``\textit{write down A on red cube}", word ``down" is tagged as ``direction'', whereas ``write down" is a single phrase referring to ``rename" action.     
\textbf{(3) Wrong ASC matching}: In user command ``\textit{move the green block to first row}", \textit{first row} is not a coordinate and thus not groundable. However, due to a small similarity with action ASC of AID 2 [see Table 2], the command is wrongly reduced to that. %``\textit{move block\_set/X1 to first row}", which can match . 
\textbf{(4) implicitly parameterized language expressions:} Considering the user command ``\textit{Enlarge paragraph 1}", the phrase ``\textit{enlarge}" is implicitly parameterized, which is equivalent to ``\textit{set font size to large}" and is difficult to ground to action ASC with AID 7 for Webpage Design (see Supplementary).

\section{Conclusion}
This paper proposed an natural language to natural language (NL2NL) approach to building natural language interfaces and a system CML, which are very different from traditional approaches. Due to the NL2NL approach, CML is application independent except that the ASCs (API seed commands) need to be specified by the application developer. Our evaluation using two applications show that CML is effective and highly promising. In our future work, apart from improving CML's accuracy, we will study how the composition of actions can be handled as well for more complex applications (e.g., database querying) and how to learn referential expressions through dialogues.

\section*{Acknowledgements}
This work was partially supported by a research gift from Northrop Grumman Corporation. 

\bibliography{emnlp-ijcnlp-2019}
\bibliographystyle{acl_natbib}

\appendix

\section{Webpage Design}
\vspace{0.1cm}
The goal of the Webpage Design application is to provide functions to the end user to help them design a web page. Our current environment deals with various html objects like image, title, paragraph, button. Action functions (APIs) are designed to manipulate these objects in the 2D space.

\subsection{Properties and Domains}
Table 1 shows different properties of an object or action (specified in the bracket next to the name of the property in the property column). The domain column in Table 1 lists the value domain of each property. E.g., ``\textit{color}" is a property of an html object, which can take five values \{’red’, ’green’, ’brown’, ’blue’, ’black’\}. Similarly, \textit{type} denotes html object type and can be one of four categories:
\{’image’, ’button’, ’title’, ’paragraph'\}. The \textit{text} written on an element like title or paragraph is denoted by double quote (`` '') in the command. The \textit{name} of an element is denoted by its type followed by an id number (like title 1, image 2, etc.) and/or some string followed by image extension for image objects (like \textit{myphoto.jpg})  in the command. The \textit{location} of an object is specified by an 2D co-ordinate ($x$, $y$) and so on. Besides these, `\textit{direction}' is a application-specific property with domain \{`\textit{left}', `\textit{right}', `\textit{above}', `\textit{below}'\} needed to execute an action for moving an object in a given direction. 

\begin{table}[t!]
\small
\centering
\caption{\small Properties and their domains for Webpage Design}
\begin{tabular}{|l|l|}
\hline
\multicolumn{1}{|c|}{\textbf{Property}} & \multicolumn{1}{c|}{\textbf{Domain}}                                                                 \\ \hline
color  (object)                                 & \begin{tabular}[c]{@{}l@{}}'red', 'green', 'brown',\\  'blue', 'black'\end{tabular}                \\ \hline
type (object)                                & \begin{tabular}[c]{@{}l@{}}'image', 'button', 'title',\\ 'paragraph'\end{tabular} \\ \hline
location (object)                                 & (x, y) coordinate in 2D space                                                                        \\ \hline
name (object)                                    & 
\begin{tabular}[c]{@{}l@{}}object's type followed by a \\number, (like image 1, title 2 \\etc.)  or xyz.jpeg, xyz.png,\\ where xyz is a string \end{tabular} 
                                                                          \\ \hline
text (object)                                    & string in quote ( " .... ")                                   \\ \hline
font\_size  (object)                                 & \begin{tabular}[c]{@{}l@{}}'small',  'medium', 'large'\end{tabular}                \\ \hline
graphics\_size  (object)                                 & \begin{tabular}[c]{@{}l@{}}'height', 'width'\end{tabular}                \\ \hline
direction (action)                              & 'right', 'left', 'above', 'below'                                                                    \\ \hline
\end{tabular}
\end{table}

\begin{table*}[t!]
\small
\centering
\caption{\small Action ASC specifications for Webpage Design and groundable example commands from user. (*) denotes that the
variable do not take part in command reduction (Utility Constraint), which is automatically identified by CML. }
\vspace{0.2cm}
\begin{tabular}{|l|c|l|l|l|}
\hline
\multicolumn{1}{|c|}{\textbf{API Function}}                                & \multicolumn{1}{l|}{\textbf{AID}} & \multicolumn{1}{c|}{\textbf{Action ASCs}}                                & \multicolumn{1}{c|}{\textbf{Variable/Argument: Type}}                                                             & \multicolumn{1}{c|}{\textbf{Example User Commands}}                                                         \\ \hline
Add(X1, X2)                                                            & \textbf{1}                        & \begin{tabular}[c]{@{}l@{}}add X1 at \\ location X2\end{tabular}             & \begin{tabular}[c]{@{}l@{}}'X1': 'type'(*), \\ 'X2': 'location'(*)\end{tabular}                                & \begin{tabular}[c]{@{}l@{}}add a title at (20, 30);\\ add an image at (30, 40)\end{tabular}                     \\ \hline
Write(X1, X2)                                                          & \textbf{2}                        & write text X1 on X2                                                          & \begin{tabular}[c]{@{}l@{}}'X1': 'text'(*), \\ 'X2': 'element\_set'\end{tabular}                            & write "My Home Page" on title 1                                                                                 \\ \hline
Remove(X1)                                                             & \textbf{3}                        & remove X1                                                                    & 'X1': 'element\_set'                                                                                     & \begin{tabular}[c]{@{}l@{}}delete title 1\\ remove image photo.png\end{tabular}                                 \\ \hline
Move(X1, X2)                                                           & \textbf{4}                        & \begin{tabular}[c]{@{}l@{}}move X1 to\\  location X2\end{tabular}            & \begin{tabular}[c]{@{}l@{}}'X1': 'element\_set', \\ 'X2': 'location'(*)\end{tabular}                        & move title 1 to (20, 30)                                                                                        \\ \hline
\begin{tabular}[c]{@{}l@{}}MoveByUnits(X1, \\ X2, X3)\end{tabular}     & \textbf{5}                        & \begin{tabular}[c]{@{}l@{}}move X1 \\ along X2 by X3 \\ units\end{tabular}   & \begin{tabular}[c]{@{}l@{}}'X1': 'element\_set', \\ 'X2': 'direction', \\ 'X3': 'number'\end{tabular}    & move image 1 left by 10 units                                                                                   \\ \hline
UpdateColor(X1, X2)                                                    & \textbf{6}                        & \begin{tabular}[c]{@{}l@{}}set color of \\ X1 to X2\end{tabular}             & \begin{tabular}[c]{@{}l@{}}'X1': 'element\_set', \\ 'X2': 'color'(*)\end{tabular}                           & \begin{tabular}[c]{@{}l@{}}color paragraph 1 as red;\\ change color of title 1 to blue\end{tabular}             \\ \hline
UpdateFont(X1, X2)                                                     & \textbf{7}                        & \begin{tabular}[c]{@{}l@{}}set font size of \\ X1 to X2\end{tabular}         & \begin{tabular}[c]{@{}l@{}}'X1': 'element\_set', \\ 'X2': 'font\_size'(*)\end{tabular}                      & make title 1 large                                                                                              \\ \hline
\begin{tabular}[c]{@{}l@{}}SetGraphicsSize(X1, \\ X2, X3)\end{tabular} & \textbf{8}                        & \begin{tabular}[c]{@{}l@{}}set the X1 of X2\\  to X3\end{tabular}            & \begin{tabular}[c]{@{}l@{}}'X1': 'graphics\_size'(*),\\ 'X2': 'element\_set',\\ 'X3': 'number'(*)\end{tabular} & \begin{tabular}[c]{@{}l@{}}set the height of image 1 to 30\\ set the width of paragraph 1 \\ to 40\end{tabular} \\ \hline
IncreaseSize(X1, X2, X3)                                               & \textbf{9}                        & \begin{tabular}[c]{@{}l@{}}increase the X1 of \\ X2 by X3 units\end{tabular} & \begin{tabular}[c]{@{}l@{}}'X1': 'graphics\_size'(*),\\ 'X2': 'element\_set',\\ 'X3': 'number'(*)\end{tabular} & \begin{tabular}[c]{@{}l@{}}increase the height of image 1\\ by 10 units\end{tabular}                            \\ \hline
DecreaseSize(X1, X2, X3)                                               & \textbf{10}                       & \begin{tabular}[c]{@{}l@{}}decrease the X1 of\\ X2 by X3 units\end{tabular}  & \begin{tabular}[c]{@{}l@{}}'X1': 'graphics\_size'(*),\\ 'X2': 'element\_set',\\ 'X3': 'number'(*)\end{tabular} & \begin{tabular}[c]{@{}l@{}}reduce the width of image 1\\ by 5 units\end{tabular}                                \\ \hline
\end{tabular}
\end{table*}

\begin{table*}[t!]
\small
\centering
\caption{\small Utility ASC specifications for Webpage Design and groundable example command sub-expressions (O for output).}
\vspace{0.2cm}
\begin{tabular}{|l|c|l|l|c|}
\hline
\multicolumn{1}{|c|}{\textbf{Function}}                                     & \textbf{AID} & \multicolumn{1}{c|}{\textbf{Utility ASCs}}                        & \multicolumn{1}{c|}{\textbf{Argument: Type}}                                                       & \textbf{Example sub-expressions}                                                                        \\ \hline
GetElementbyLocation(X1)                                                    & \textbf{11}  & location/X1                                                           & \begin{tabular}[c]{@{}l@{}}X1: 'location', \\ O1: 'element\_set'\end{tabular}                      & element at (20, 30)                                                                                  \\ \hline
GetElementbyType(X1)                                                        & \textbf{12}  & X1                                                                    & \begin{tabular}[c]{@{}l@{}}X1: 'type',\\ O1: 'element\_set'\end{tabular}                           & \begin{tabular}[c]{@{}c@{}}get all titles; \\ get all paragraphs\end{tabular}                        \\ \hline
GetElementbyFont(X1)                                                        & \textbf{13}  & X1                                                                    & \begin{tabular}[c]{@{}l@{}}X1: 'font\_size',\\ O1: 'element\_set'\end{tabular}                     & \begin{tabular}[c]{@{}c@{}}elements having \\ size large\end{tabular}                                \\ \hline
\begin{tabular}[c]{@{}l@{}}GetElementby\\ GraphicsSize(X1, X2)\end{tabular} & \textbf{14}  & X1 of X2                                                              & \begin{tabular}[c]{@{}l@{}}X1: 'graphics\_size',\\ X2: 'number',\\ O1: 'element\_set'\end{tabular} & \begin{tabular}[c]{@{}c@{}}elements having\\ height of 20\end{tabular}                               \\ \hline
GetElementbyColor(X1)                                                       & \textbf{15}  & X1                                                                    & \begin{tabular}[c]{@{}l@{}}X1: 'color',\\ O1: 'element\_set'\end{tabular}                          & \begin{tabular}[c]{@{}c@{}}red element;\\   blue title\end{tabular}                                  \\ \hline
GetElementbyText(X1)                                                        & \textbf{16}  & X1                                                                    & \begin{tabular}[c]{@{}l@{}}X1: 'text',\\ O1: 'element\_set'\end{tabular}                           & \begin{tabular}[c]{@{}c@{}}element with text\\ "welcome!"\end{tabular}                               \\ \hline
GetLocation(X1, X2)                                                         & \textbf{17}  & \begin{tabular}[c]{@{}l@{}}get location\\ along X1 of X2\end{tabular} & \begin{tabular}[c]{@{}l@{}}X1: 'direction',\\ X2: 'element\_set',\\ O1: 'location'\end{tabular}    & \begin{tabular}[c]{@{}c@{}}location at the left of\\ image 2; location \\ below title 1\end{tabular} \\ \hline
GetElementbyName(X1)                                                        & \textbf{18}  & X1                                                                    & \begin{tabular}[c]{@{}l@{}}X1: 'name',\\ O1: 'element\_set'\end{tabular}                           & \begin{tabular}[c]{@{}c@{}}title 1; image 2;\\ profile.jpeg\end{tabular}                             \\ \hline
\end{tabular}
\end{table*}

\subsection{ASC Specification}
Table 2 shows the action ASC (API seed command) and Table 3 shows the utility ASC specifications for Webpage Design. Some example user commands or sub-expressions that can fire them are also given. Here, the argument type `element\_set' denotes a set of unique html (instantiated) object ids. We define 10 action ASCs and 8 utility ASCs (for their respective APIs) to ground a user command. The action ASCs are: adding an html object at location (x,y) [AID 1], writing some text on an element like title or paragraph [AID 2], removing an element [AID 3], moving elements [AID 4-5], changing  properties of an object [AID 6-10]. Utility ASCs either retrieve the element set having a given property and value [AID 11-16, 18] or return a location object ($x$, $y$) in a given \textit{direction} of an object [AID 17]. 

\end{document}